\documentclass[runningheads]{llncs}
\usepackage{graphicx}
\usepackage{tikz}
\usepackage{comment}
\usepackage{amsmath,amssymb} % define this before the line numbering.
\usepackage{color}
\usepackage{multirow}
\usepackage{mathrsfs}
\usepackage{subfig}

\usepackage{makecell}
\newcommand{\PreserveBackslash}[1]{\let\temp=\\#1\let\\=\temp}
\newcolumntype{C}[1]{>{\PreserveBackslash\centering}p{#1}}
\newcolumntype{R}[1]{>{\PreserveBackslash\raggedleft}p{#1}}
\newcolumntype{L}[1]{>{\PreserveBackslash\raggedright}p{#1}}

\begin{document}
% \linenumbers
\pagestyle{headings}
\mainmatter
\def\ECCVSubNumber{5061}  % Insert your submission number here

\title{Temporal Complementary Learning for Video Person Re-Identification} % Replace with your title

\begin{comment}
\titlerunning{ECCV-20 submission ID \ECCVSubNumber} 
\authorrunning{ECCV-20 submission ID \ECCVSubNumber} 
\author{Anonymous ECCV submission}
\institute{Paper ID \ECCVSubNumber}
\end{comment}
%******************

%\begin{comment}
\titlerunning{Temporal Complementary Learning for Video Person Re-Identification}
\author{Ruibing Hou\inst{1,2} \and
Hong Chang\inst{1,2} \and
Bingpeng Ma\inst{2} \and
Shiguang Shan\inst{1,2,3} \and
Xilin Chen\inst{1,2}
}
\authorrunning{R. Hou, H. Chang et al.}
\institute{Key Laboratory of Intelligent Information Processing of Chinese Academy of Sciences (CAS), Institute of Computing Technology, CAS, Beijing 100190,  China \and
University of Chinese Academy of Sciences, Beijing 100049, China \and
CAS Center for Excellence in Brain Science and Intelligence Technology, China\\
\email{ruibing.hou@vipl.ict.ac.cn,  \{changhong, sgshan,xlchen\}@ict.ac.cn, bpma@ucas.ac.cn}
}
%\end{comment}
%******************
\maketitle

\begin{abstract}
This paper proposes a \textit{Temporal Complementary Learning Network} that extracts complementary features of consecutive video frames for video person re-identification. Firstly, we introduce a  \textit{Temporal Saliency Erasing} (TSE) module including a saliency erasing operation and a series of ordered learners. Specifically, for a specific frame of a video, the saliency erasing operation drives the specific learner to mine new and complementary parts by erasing the parts activated by previous frames. Such that the diverse visual features can be discovered for consecutive frames and finally form an integral characteristic of the target identity. Furthermore, a \textit{Temporal Saliency Boosting} (TSB) module is designed to propagate the salient information among video frames to enhance the salient feature. It is complementary to TSE by effectively alleviating the information loss caused by the erasing operation of TSE. Extensive experiments show our method performs favorably against state-of-the-arts. The source code is available at \url{https://github.com/blue-blue272/VideoReID-TCLNet}.

\end{abstract}

\section{Introduction}
Person re-identification (reID) aims at retrieving particular persons from non-overlapping camera views. It plays a significant role in video surveillance analysis. Image person reID has achieved great progress in term of the methods~\cite{PCB,IANet,zhang2019densely} and large benchmarks construction~\cite{Market1501,Duke,msmt17}. Recently, with the emergence of large video benchmarks~\cite{mars,dukevideo} and the growth of computational resource, video person reID has been attracting a significant amount of attention. The video data contain richer spatial appearance information and temporal cues, which can be exploited for more robust reID.

Current state-of-the-art approaches for video person reID are based on deep neural networks. The mainstream approaches usually consider a sequence of frames as input, and utilize convolutional neural networks (CNNs) to extract the feature for each frame independently, followed by temporal feature aggregation, \textit{e.g.}, through recurrent layer~\cite{RCN,jointly,See} or temporal attention layer~\cite{QAN,RQAN,jointly}. With the powerful deep networks and large-scale labeled benchmarks, these methods achieve favorable performance and efficiency.

\begin{figure}[t]
\begin{center}
\centerline{\includegraphics[width=0.7\textwidth]{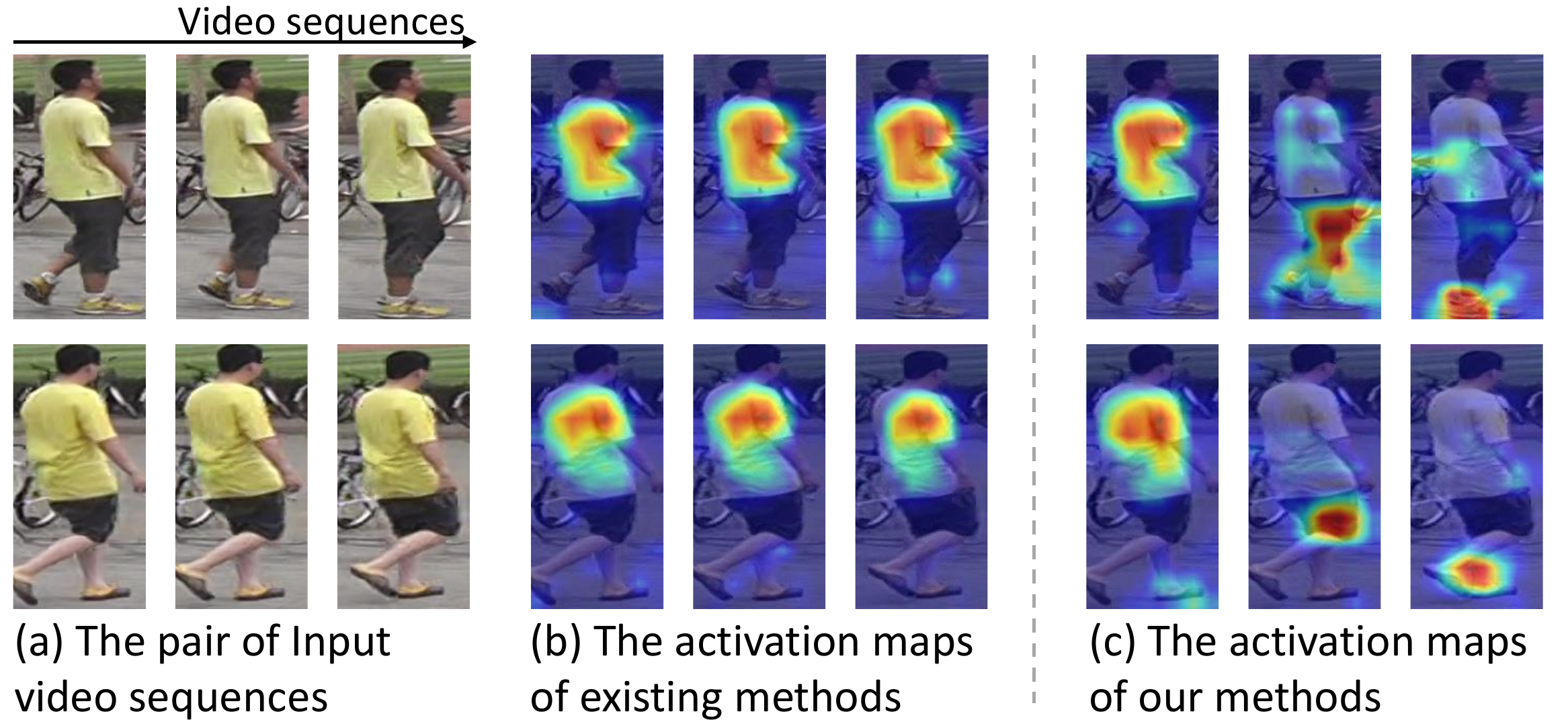}}
\caption{An example of class activation maps~\cite{zhou2016learning} of a pair of input video sequences of existing method~\cite{QAN} and our method. Warmer color with higher value}
\label{motivation}
\end{center}
\end{figure}

Despite the significant progress in video person reID, most existing methods do not take full advantage of the rich spatial-temporal clues in the video. To be specific, since the pedestrian frames of a video are highly similar and the existing methods perform the same operation on each frame, these methods typically produce highly redundant features for the frames of a video. The redundant features typically attend to the same local salient parts~\cite{hou2019cross}, which are difficult to distinguish the persons with similar appearance.  For example, as illustrated in Fig.~\ref{motivation}(b), the upper clothes of the sequence pair attract the most attention, but are difficult to distinguish the two pedestrians. Therefore, it is appealing to explore a way of fully mining the spatial-temporal clues in the video, which can discover diverse visual cues for different frames of a video to form a full characteristic of each identify.

In this paper, we propose a \textit{Temporal Complementary Learning Network} (TCLNet) for fully exploiting the spatial-temporal information of the video data. Firstly,  we introduce a \textit{Temporal Saliency Erasing} (TSE) module consisting of a saliency erasing operation and a series of ordered adversary learners. The key idea is to extract complementary features for consecutive frames of a video by the adversary learners. In particular, the first learner is firstly leveraged to extract the most salient feature for the first frame of a video sequence. Then, for the feature map of the second frame, the \textit{saliency erasing operation} utilizes the \textit{temporal cues} to erase the region attended by the first learner. Then we feed the feature excluding the erased region into the second learner for discovering new and complementary parts. By recursively erasing all previous discovered parts, the ordered learners can mine complementary parts for the consecutive frames and finally obtain an integral characteristic of the target person. As shown in Fig.~\ref{motivation}(c), with TSE, the features of  consecutive frames can focus on diverse parts, covering the whole body of the target identity. 

However, as TSE recursively erases the most salient part of the second and subsequent frames of the input video sequence, the representation of the most salient part is less powerful.  To this end, we propose a \textit{Temporal Saliency Boosting} (TSB) module to enhance the representational power of the most salient part. Concretely, TSB utilizes the \textit{temporal cues} to propagate the most salient information among the video frames. In this way, the most salient features can capture the visual cues across all frames of video, hence exhibit stronger representational capability.

The proposed two modules can be inserted in any deep CNNs to extract complementary features for video frames. To the best of our knowledge, it is the first attempt to extract complementary features for the consecutive frames of a video. We demonstrate the effectiveness of the proposed method on three challenging video reID benchmarks, and our method outperforms the state-of-the-art methods under multiple evaluation metrics.

\section{Related Work}
\textbf{Person Re-identification.} Person reID for still images has been extensively studied~\cite{PCB,ge2018deep,zhang2019densely,zheng2019re,CRF,liu2019view,IANet,gu2019temporal}. Recently, researchers start to pay attention to video reID~\cite{QAN,diversity,jointly,VRSTC,Gu3D}. The existing methods can be divided into two categories, \textit{i.e.}, image-set based methods and  temporal-sequence based methods.

Image-set based methods consider a video as a set of disordered images. These methods usually extract the features of each frame independently, then use a specific temporal pooling strategy to aggregate the frame-level features. For example,  Zheng \textit{et al.}~\cite{mars} apply an average pooling across all frames to obtain the video feature. The works~\cite{QAN,RQAN,diversity,AD-zhao} further use a temporal attention mechanism that assigns a quality score to each frame for weighted average pooling. 
These methods exhibit promising efficiency, but totally ignore the temporal cues of the video data. 

Temporal-sequence based methods exploit the temporal cues for video representation learning. The early works~\cite{RCN,jointly,See} use the optical flows to encode the short-term motion information among adjacent frames. Mclaughlin~\textsl{et al.}~\cite{wang2017fast} propose a recurrent architecture~\cite{hou2019video} to aggregate the frame-level representations and  yield a sequence-level feature representations. Zhang~\textsl{et al.}~\cite{zhang2020ordered} argue that the recurrent structure may not be optimal to learn temporal dependencies and propose to learn orderless ensemble ranking. Liao~\textsl{et al.}~\cite{V3DP} propose to use 3D convolution for spatial-temporal feature learning. Recently, some works~\cite{V3DP,VRSTC,GLTL} apply non-local blocks~\cite{non-local} to capture long-term temporal cues.  

Nevertheless, these methods perform the same operation on each frame, leading to that the features of different frames are highly redundant. Therefore in this work, we propose a temporal complementary learning mechanism to extract complementary features for consecutive frames of a video, which is able to obtain an integral characteristic of the target identity for better reID.

\textbf{Erasing Pixels or Activations.} \textit{Image based erasing} has been widely applied as a data augmentation technique. For instance, the works~\cite{zhong2017random,devries2017improved} randomly erase a rectangle region of the input images during training. Singh \textit{et al.}~\cite{singh2017hide} propose to divide the image into a grid with fixed patch size and randomly mask each patch.
\textit{Feature based erasing} typically drops the feature activations. Dropout~\cite{srivastava2014dropout} drops the feature units randomly, which is a widely used regularization technique to prevent overfitting. DropBlock~\cite{ghiasi2018dropblock} randomly drops a contiguous region of the convolutional features for CNNs. In the work \cite{dai2018batch}, all feature maps in the same batch are dropped in a consistent way for better metric learning. 

Our work is fundamentally different from the existing erasing methods in two folds. Firstly, \textbf{the basic idea is different}. The above methods typically use the erasing strategy to regularize the \textit{training} of networks to  prevents overfitting. Differently, our method uses the erasing operation to extract complementary features for video frames during both training and testing phases. Secondly, \textbf{the erasing mechanism is different.} The existing methods usually randomly erase the pixels or activations without any high-level guidance. Our method erases the regions for each frame guided by the activated parts of previous frames, which guarantees the frame pays attention to new person parts thus is more efficient. 

Recently, Liu~\textit{et al.}~\cite{liu2019pose,liu2020adversarial} also propose to use saliency to guide erasing to learn complementary features. However, the purpose and implementation of these methods are different from ours. Specifically, these methods perform \textit{image erasing} on \textit{a singe image input}. On the contrary, our method performs \textit{feature erasing} on \textit{a video sequence} that allows for extracting complementary features for consecutive video frames. 

\section{Temporal Complementary Learning Network}
The proposed Temporal Complementary Learning Network for video person reID includes two novel components, \textit{i.e.} TSE for complementary feature mining and TSB for enhancing the most salient feature.

\subsection{Temporal Saliency Erasing Module}
Compared to images, video data contain richer spatial-temporal information, which should be fully exploited for more robust feature representation. However, most existing methods perform the same operation on each frame, resulting in highly redundant features of different frames that only highlight a local part~\cite{zheng2019re,CAMA}, as shown in Fig.~\ref{motivation} (b). To this end, we design a \textit{temporal saliency erasing module} to mine complementary parts from consecutive frames of a video to form an integral characteristic of the target person.

\begin{figure}[t]
\begin{center}
\centerline{\includegraphics[width=1.0\textwidth]{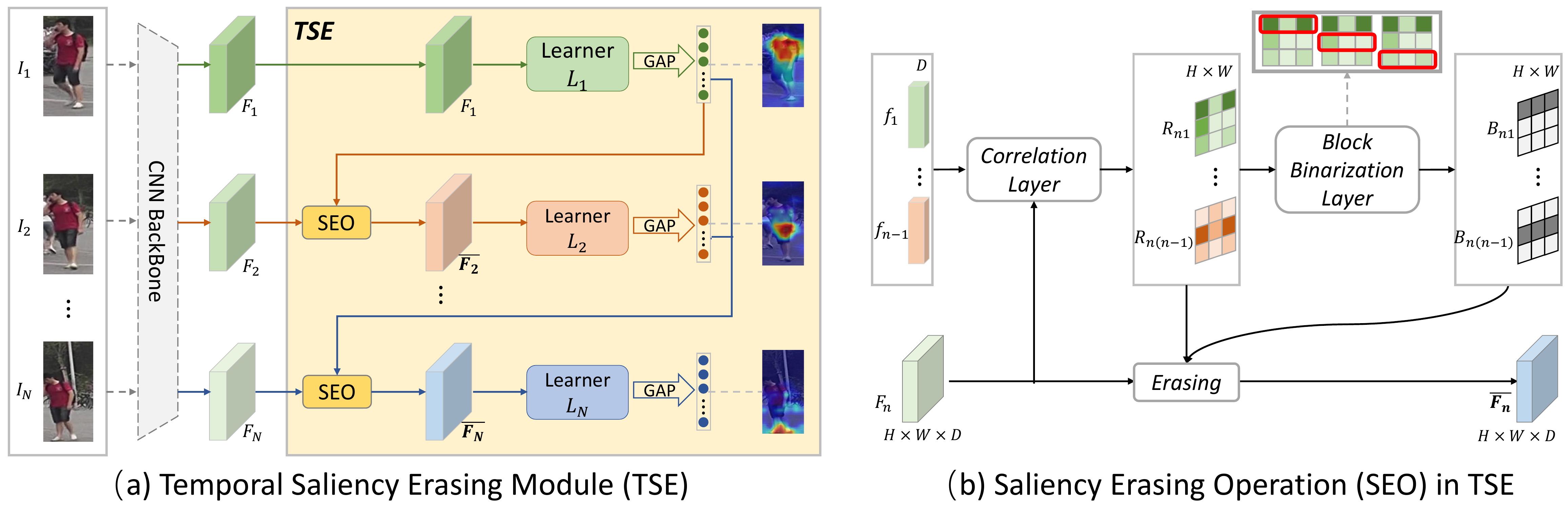}}
\caption{(a) The architecture of TSE. (b) \textit{Saliency erasing operation} in TSE for feature map $F_n$ ($n>1$)}
\label{TSE}
\end{center}
\end{figure}

\textbf{TSE Overview.} As shown in Fig.~\ref{TSE} (a), TSE iteratively performs two operations on each frame: adversarially erasing the parts discovered by previous frames with a \textit{saliency erasing operation} (SEO) and learning a specific \textit{learner} for discovering new part to extract complementary feature. Concretely, the input of TSE is the set of frame-level feature maps $\{F_n\}_{n=1}^N (F_n \in\mathbb{R}^{H\times W \times D}$) of a video segment $\{I_n\}_{n=1}^N$ after a CNN Backbone, where the video segment contains $N$ consecutive frames and $n$ is the index of the video frame, and $H$, $W$ and $D$ denote the height, width and channel number of the feature map respectively. Firstly, TSE uses the learner $L_1$ followed by a GAP (global average pooling) layer to extract the most salient feature $f_1\in\mathbb{R}^{D_1}$ for $I_1$, which is denoted as:
\begin{equation}
 f_1=\text{\text{GAP}}(L_1(F_1)).
\label{eq1}
\end{equation}
Then, SEO erases the feature map $F_2$ of frame $I_2$ guided by the mined discriminative part of $f_1$. The erased feature map is then fed into learner $L_2$. As the part attended by learner $L_1$ has been removed, the learner $L_2$ is naturally driven to discover new discriminative parts for identifying the target person. Recursively, for the frame feature map $F_n \left(n>1\right)$, TSE firstly applies SEO to erases all parts discovered by previous frames to form the erased feature map $\overline{F_n}$, and then uses its specific learner $L_n$ to mine new parts and obtain the feature vector $f_n$, which can be formulated as:
\begin{equation}
\overline{F_n} = \text{\text{SEO}}(F_n;f_1, \dots, f_{n-1}), \ \  f_n=\text{\text{GAP}}(L_n(\overline{F_n})) \ (\ 1< n\leq N).
\label{eq2}
\end{equation}
The \textit{saliency erasing operation} and \textit{learners} repeatedly perform on the $N$ consecutive frames of the input segment. Finally, the integral characteristic of target person can be obtained by combining the features produced by these frames.

\textbf{Correlation Layer of SEO.} The implementation of SEO (Eq.~\ref{eq2}) is illustrated in Fig.~\ref{TSE} (b). For the feature map $F_n$ of frame $I_n$ to be erased, a \textit{correlation layer} is firstly designed to obtain the correlation maps between previous-frame feature vectors $f_k (k < n)$ and $F_n$.  In particular, we firstly consider the feature vector at every spatial location $(i,j)$ of $F_n$ as a $D$ dimensional local descriptor  $F_n^{\left(i,j\right)}$. Then the \textit{correlation layer} computes the semantic relevance between $f_k$ and all the local descriptors of $F_n$ with dot-produce similarity~\cite{non-local} to get the corresponding correlation map $R_{nk}\in\mathbb{R}^{H\times W}$ as:
\begin{equation}
R_{nk}^{(i,j)} = (F_n^{(i,j)})^T \left(w^T f_k\right) \quad (1\leq i \leq H, 1\leq j \leq W, 1\leq k \leq n-1).
\label{eq3}
\end{equation}
Here $w\in\mathbb{R}^{D_1\times D}$ projects $f_k$ to the feature space of $F_n$, matching the number of channels to that of $F_n$.  Eq.~\ref{eq3} shows that the local descriptors that describe the part activated by $f_k$ tend to present higher relevance values in $R_{nk}$. Thus $R_{nk}$ can localize the regions in $F_n$ of the parts activated by the previous-frame feature vector $f_k$. 

\textbf{Block Binarization Layer of SEO.} The correlation maps are then used to generate the binary masks to identify the regions to be erased. A valiant approach is to conduct a threshold on the correlation maps. However, it usually produces noncontinuous regions. As pointed by~\cite{ghiasi2018dropblock}, since the convolutional feature units are correlated spatially, when erasing the feature units discontinuously, information about the erased units can still be transmitted to the next layer. To this end, we design a \textit{block binarization layer} to generate the binary mask which can erase a contiguous region of a feature map. As shown in Fig.~\ref{TSE} (b), we search the most highlighted continuous area in the correlation map using a \textit{sliding block}. Formally, for a correlation map of size $H\times W$ and the sliding block of size $h_e\times w_e$, when we move the block with horizontal and vertical strides $s_w$ and $s_h$ respectively, the total number of block position can be computed as $N_{pos}=\left(\lfloor\frac{H-h_e}{s_h}\rfloor + 1\right) \times \left(\lfloor\frac{W-w_e}{s_w}\rfloor + 1\right)$. Thus $N_{pos}$ candidate blocks can be obtained for each correlation map. We then define the correlation value of a block as the sum of the correlation values of the items in the block. Finally, we select the candidate block with the highest correlation value as the block to be erased, \textsl{i.e.}, the binary mask $B_{nk}\in\mathbb{R}^{H\times W}$ of correlation map $R_{nk} $ is generated by setting the values of the units in the selected block to $0$ and others to $1$. We then merge the masks $\{B_{nk}\}_{k=1}^{n-1}$ to a fused mask $B_n$ for the feature map $F_n$, which is calculated as:
\begin{equation}
B_n = B_{n1} \odot B_{n2} \odot \dots \odot B_{n(n-1)},
\end{equation}
where $\odot$ is element-wise product operation.

\textbf{Erasing Operation in SEO.}  In order to make TSE end-to-end trainable, we employ a gate mechanism to erase the feature map $F_n$. In particular, we apply a softmax layer to the fused correlation map, from which we erase the selected block using $B_n$ to obtain a gate map $G_n\in\mathbb{R}^{H\times W}$:
\begin{equation}
G_n = \text{softmax}\left(R_{n1} \odot R_{n2} \odot \dots \odot R_{n(n-1)}\right)  \odot B_n.
\end{equation}
$F_n$ is erased based on $B_n$ and $G_n$ to generate the erased feature map $\overline{F_n}$.
For consistency, we also apply the erasing operation on $F_1$ with a binary mask filling with 1. In this way, the gradients can propagate to the parameter $w$ of SEO and TSE can be trained by back-propagation.

\subsection{Temporal Saliency Boosting Module}

Although TSE can extract complementary features for input segment frames, the \textit{saliency erasing operation} inevitably leads to information loss of the most salient part.
To address this problem, we propose a \textit{temporal saliency boosting module}. Since before the erasing operation, the intermediate feature maps of all high quality frames usually focus on the most salient parts. TSB is proposed to propagate the most salient information among the intermediate frame-level feature maps. In this way, the most salient feature can fully capture the visual cues in all frames thus presents strong discriminative power.

\begin{figure}[t]
\begin{center}
%includegraphics[height=6.5cm]{eijkel2}
\centerline{\includegraphics[width=0.95\textwidth]{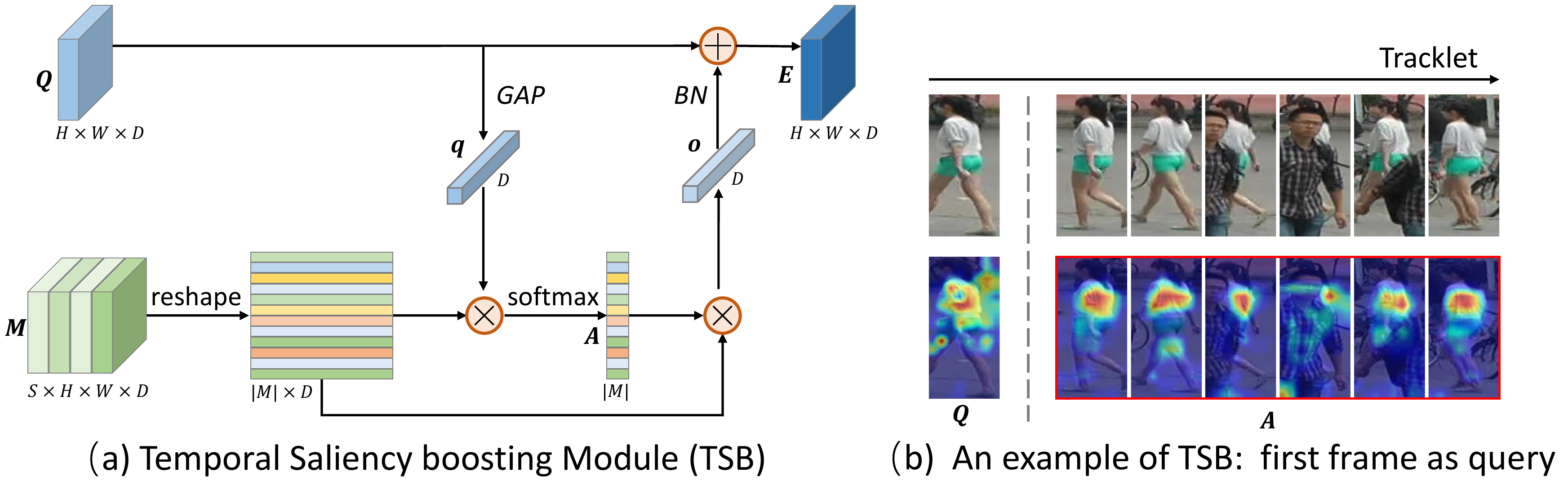}}
\caption{(a) The architecture of TSB. (b) Visualization of probability map $A$ computed on a video sequence, where we take the first frame as the query}
\label{TSB}
\end{center}
\end{figure}

The structure of TSB is illustrated in Fig.~\ref{TSB} (a), which takes a set of frame-level feature maps of a video as the input. TSB is based on a \textit{query-memory} attention mechanism, where we respectively consider the feature map of each frame as the query $Q\in\mathbb{R}^{H\times W\times D}$, and the memory $M\in\mathbb{R}^{S\times H\times W\times D}$ containing a collection of feature maps from the remaining $S$ frames of the video is used to enhance the representational power of the query. Specifically, we firstly squeeze $Q$ to a descriptor which can describe the query statistics. This is achieved by using GAP (global average pooling) to generate the channel-wise statistics $q\in\mathbb{R}^{D}$. Then we reshape $M$ to $\mathbb{R}^{|M|\times D}$ ($|M|=S\times H \times W$) which can be viewed as a set of $D$-dimensional local descriptors. Corresponding, a probability map $A \in\mathbb{R}^{|M|}$ can be obtained regarding how well the query vector matches each descriptor of the memory through cosine similarity:
\begin{equation}
A_{i} = \frac{\exp(\tau \bar{q}^T \bar{M}_{i})}{\sum_{j=1}^{|M|} \exp(\tau \bar{q}^T \bar{M}_{j})},
\label{eq7}
\end{equation}
where $M_i\in\mathbb{R}^D$ denotes the $i^{th}$ local descriptor of $M$, $\bar{q}$ and $\bar{M}_{i}$ is the normalized $q$ and $M_i$ with $L_2$ norm, and $\tau$ is the temperature hyperparameter. The output $o\in\mathbb{R}^{D}$ is then calculated as the sum of all the items in the memory weighted by their probabilities as $o=M^TA$. In this way, the descriptors of $M$ that are similar to the query present higher weights, which can avoid the corruption of low quality frames, \textit{e.g.}, the occluded frames shown in Fig.~\ref{TSB} (b).

At last, we propagate the weighted descriptor $o$ to the query $Q$ with a residual learning scheme, which is defined as $E=\text{BN}(o)+Q$. Here, BN is a batch normalization~\cite{BN} layer to adjust the scale of $o$ to the query $Q$. Notably, before entering the BN layer, $o\in\mathbb{R}^{D}$ is duplicated along the spatial dimensions to $\mathbb{R}^{H\times W\times D}$ to be compatible with the size of $Q$.

Taking the first frame of a video sequence as the query example, Fig.~\ref{TSB} (b) visualizes its initial feature map $Q$ and corresponding probability map $A$. We can observe that $Q$ roughly focuses on the most salient part (\textit{i.e.}, shirts), and the probability map $A$ can localize this part in other frames. Notably, through the \textit{query-memory} matching, the corrupted frames presents lower weights in $A$, which indicates their features are suppressed during propagation and the output feature is robust to corruption. With the information propagation, the output feature fully captures the most salient visual cues in all frames of the video thus presents stronger representational power.

\subsection{Overall Architecture}

\begin{figure}[t]
\begin{center}
%includegraphics[height=6.5cm]{eijkel2}
\centerline{\includegraphics[width=0.95\textwidth]{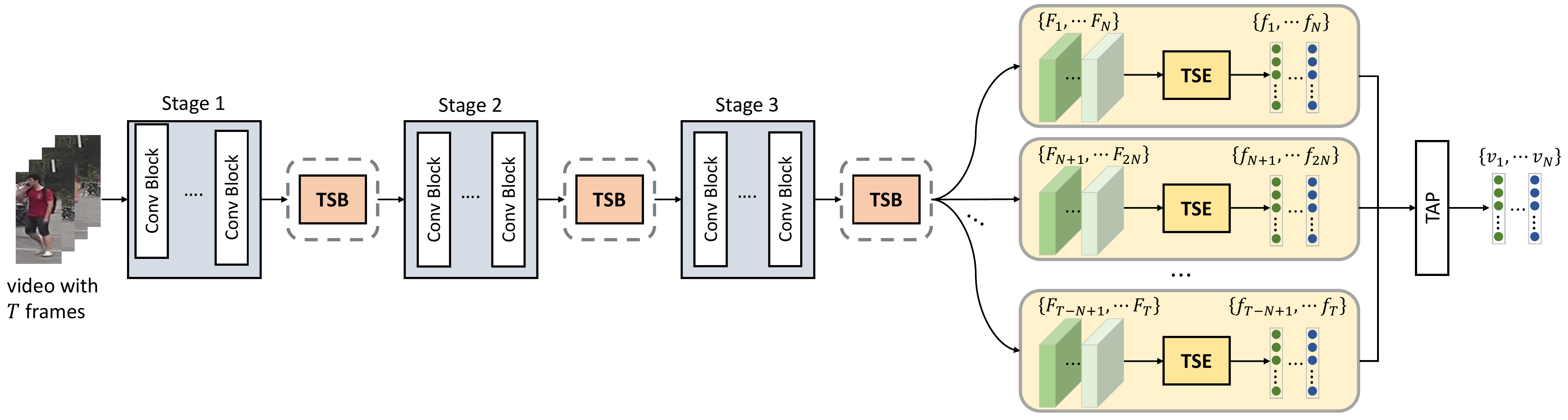}}
\caption{The architecture of TCLNet. TAP denotes temporal average pooling layer}
\label{network}
\end{center}
\end{figure}

The architecture of Temporal Complementary Learning Network (TCLNet) that integrates TSE and TSB modules is illustrated in Fig.~\ref{network}. Our network is built on ResNet-50~\cite{residual} pretrained on ImageNet~\cite{imagenet}. ResNet-50 consists of four consecutive stages, \textit{i.e.}, \textit{stage}$1$$\sim$$4$, which respectively contains $3$, $4$, $6$ and $3$  residual blocks. We adopt the first three stages (\textit{stage}$1$$\sim$$3$) as the backbone and the last stage (\textit{stage}$4$) as the learners of TSE. TSB can be inserted into the backbone to any stage and TSE is added to the end of the backbone. In order to reduce the network complexity, the $N$ learners of TSE share the same parameters for the first two residual blocks and have their own parameters in the last block. 

Formally, given an video consisting of $T$ consecutive frames,  the backbone with inserted TSB firstly extracts features for each frame, which is denoted as $\mathcal{F}=\{F_1,F_2,\dots, F_T\}$. Since the number of discriminative visual cues is usually finite, we only extract complementary features for $N$ ($N < T$) consecutive frames. In particular, we equally divide $\mathcal{F}$ into $L$ segments $\{C_k\}_{k=1}^{L}$ where each segment contains $N$ consecutive feature maps, \textit{i.e.}, $C_k=\{F_{(k-1)N+1}, \dots, F_{kN}\}$. We then respectively feed each segment into TSE to extract complementary features for the segment frames:
\begin{equation}
 c_k = \{f_{(k-1)N+1}, \dots, f_{kN}\}=\text{TSE}(F_{(k-1)N+1}, \dots, F_{kN})=\text{TSE}(C_k). %(1\leq k \leq L).
 \label{eq7}
 \end{equation}
We finally apply temporal average pooling on $\{c_k\}_{k=1}^L$ that aggregates the set of segment features to generate the video features $\{v_1, \dots v_N\}$.
In the test stage, the final video representation $v$ can be obtained by concatenating the feature vectors extracted by all learners, \textit{i.e.}, $v=[\hat{v}_1, \dots, \hat{v}_N]$, where $\hat{v}_i$ denotes the L2-normalization of $v_i$.

\textbf{Objective Function.} \ Following the standard identity classification paradigm \cite{PCB,IANet}, we add a classification layer to each video vector $v_i$. Cross entropy loss is then used for every $v_i$ to guide the training of the corresponding learner. Recently, some works~\cite{Co-segmentation,V3DP} use the combination of cross entropy loss and batch triplet loss~\cite{Triplet} to train the network. To fairly compare with this methods, we also explore a batch triplet loss during training.

\section{Experiments}
\subsection{Dataset and Settings}
\textbf{Datasets.} MARS~\cite{mars}, DukeMTMC-VideoReID~\cite{dukevideo} and iLIDS-VID~\cite{ilids} datasets are used for evaluation. 

\textbf{Implementation Details.} Our method is implemented using the PyTorch framework~\cite{pytorch}. During training, we sample four frames from each video sequence as input and each frame is resized to $256\times 128$. We only adopt random flipping for data augmentation. The initial learning rate is set to $0.0003$ with a decay factor $0.1$ at every $40$ epochs. Adam optimizer~\cite{adam} is used with a mini-batch size of $32$ for $150$ epochs training. TSB is added to $\text{stage}2$ of the backbone. In TSE, the number of learners, \textit{i.e}, the number of frames in each divided segment ($N$ in Eq.~\ref{eq7}), is set to $2$, the height of erased block $h_e$ is set to $3$ and the width $w_e$ is set to $8$, and the strides $s_h$ and $s_w$ of sliding block are both set to $1$. During testing, given an input of entire video, the video feature is extracted using the trained TCLNet for retrieval under cosine distance. Notably, \textit{we use all the frames of a video to obtain the video feature in the testing phase.}

\subsection{Comparison with State-of-the-art Methods}
\begin{table}[t]
\caption{Comparison with related methods on MARS, DukeMTMC-VideoReID and iLIDS-VID datasets. The methods are separated into two groups: image-set based methods (\textbf{IS}) and temporal-sequence based methods (\textbf{TS}). * denotes those trained with the combination of cross entropy loss and triplet loss}
\small
\centering
\begin{tabular}{c | l |C{1.1cm} C{1.1cm}  |C{1cm} C{1cm} | C{2cm}}
\hline
\multicolumn{2}{c|}{\multirow{2}*{Methods}} & \multicolumn{2}{c|}{MARS}  &\multicolumn{2}{c|}{Duke-Video} &iLIDS-VID\\   
\cline{3-7}
\multicolumn{2}{c|}{ } &mAP &top-1 &mAP &top-1 &top-1  \\
\hline
\multirow{7}*{\textbf{IS}}&Mars~\cite{mars} & 49.3 & 68.3 & - &- & 53.0 \\
&SeqDecision*~\cite{sequence-decision} & - & 71.2 & - &- &60.2\\
 &QAN*~\cite{QAN} & 51.7 &73.7 &- &- & 68.0\\
&DRSA~\cite{diversity} &65.8 & 82.3 & - &- &80.2\\
&EUG~\cite{dukevideo} &67.4& 80.8 & 78.3& 83.6& - \\ 
%&TriNet~\cite{Triplet} &67.7 & 79.8 &- &- &- \\
&AttDriven~\cite{AD-zhao} &78.2&87.0 &- &- &86.3\\
\hline
\multirow{9}*{\textbf{TS}} &ASTPN~\cite{jointly} & - & 44.0 & - &- & 62.0 \\
&SeeForest*~\cite{See} & 50.7 & 70.6 &- &- & 55.2\\
&DuATM*~\cite{dual-attention}& 67.7 & 81.2 &- &- &-\\ 
&M3D~\cite{M3D} &84.4 &74.1 & & &84.4\\
&Snipped~\cite{snippet} &76.1 & 86.3 &- &- & 85.4\\
&V3D*~\cite{V3DP} &77.0 &84.3 &- &- &81.3\\
&GLTP~\cite{GLTL}  & 78.5 & 87.0 & 93.7 & \textbf{96.3} & 86.0\\
&COSAM*~\cite{Co-segmentation} &79.9 &84.9 &94.1 &95.4 &-\\
&VRSTC~\cite{VRSTC} & 82.3 & 88.5 & 93.5 & 95.0 & 83.4\\
\hline
\multirow{2}*{}
&TCLNet &\textbf{83.0} &\textbf{88.8} &\textbf{95.2} &\textbf{96.3} &84.3\\
&TCLNet-tri* &\textbf{85.1} &\textbf{89.8} &\textbf{96.2} &\textbf{96.9} &\textbf{86.6} \\
\hline
\end{tabular}
\label{tab-s}
\end{table}
In Tab.~\ref{tab-s}, we compare our method with state-of-the-arts on MARS, DukeMTMC-VideoReID and iLIDS-VID datasets. Our method outperforms the best existing methods. It is noted that: \textbf{(1)} The gaps between our results and those that consider the video as a set of unordered images~\cite{mars,sequence-decision,QAN,diversity,dukevideo,Triplet,AD-zhao} are significant: about $5\%$ mAP improvement on MARS. The significant improvements demonstrate that it is effective to employ the temporal cues for video reID. \textbf{(2)} Recent works~\cite{M3D,V3DP,VRSTC} uses 3D CNN or non-local blocks to learn the temporal cues, which require high computational complexity. Our TCLNet puts much less overheads with a better performance on MARS: about $1\%$ mAP improvement. We attribute this improvement to the complementary features learned from video frames which enhance the discriminative capability of reID models. \textbf{(3)} The works~\cite{sequence-decision,QAN,See,dual-attention,V3DP,Co-segmentation} use the triplet loss to promote the performance. To fairly compare with them, we also adopt a triplet loss which further increases our performance by about $1\%$. It still outperforms the best performing work~\cite{Co-segmentation} by a large margin.

\subsection{Ablation Study}
We investigate the effectiveness of TSE and TSB modules by conducting a series of ablation studies on MARS dataset. We adopt ResNet-50~\cite{residual} with temporal average pooling as baseline (denoted as base.). In this part, all models are trained with only cross entropy loss. 

\begin{table}[t]
%\captionsetup{font={small}}
\caption{Component analysis of the proposed network on MARS. We also report  the number of floating-point operations (GFLOPs) for a four-frames sequence, and the parameter number (Params) of the models}
\centering
\begin{tabular}{L{4.3cm} |C{1.4cm} C{1.4cm}|C{1.0cm} C{1.0cm} }
\hline 
Models  & GFLOPs &Params & mAP & top-1  \\
\hline
base. &16.246  &23.5M &79.6 &86.8  \\
\hline
base.+TSE-wo-SEO &16.246 &27.9M &80.9 &87.2 \\
base.+TSE &16.251 & 29.9M &\textbf{82.5} &\textbf{88.2} \\
\hline
base.+TSB &16.254  & 23.5M &82.3 &87.6\\
TCLNet(TSE+TSB)  &16.259  & 29.9M &\textbf{83.0} &\textbf{88.8}\\
\hline
TCLNet(TSE+TSB-stage1)  &16.267 &29.9M  &82.2 &87.4\\
TCLNet(TSE+TSB-stage2)  &16.259 &29.9M &\textbf{83.0} &\textbf{88.8}\\
TCLNet(TSE+TSB-stage3)  &16.255 &29.9M &82.6 &88.2\\
TCLNet(TSE+TSB-stage23)  &16.263 &29.9M &82.7 &88.2\\
\hline
\end{tabular}
\label{tab1}
\end{table}
\textbf{Effectiveness of TSE.} We firstly evaluate the effect of TSE by replacing the stage4 layer of baseline with TSE (\textbf{base.+TSE}). As shown in Tab.~\ref{tab1}, TSE module improves the performance remarkably. Compared with the baseline, employing TSE brings $2.9\%$ mAP and $1.4\%$ top-1 accuracy gains respectively with negligible computational overhead. We argue that the learners of TSE work collaboratively to mine complementary parts so as to generate integral characteristic of the target identity, which helps to distinguish different identities with seemingly similar local parts. An example is shown in Fig.~\ref{vis1}.

\textbf{Effectiveness of \textit{Saliency Erasing Operation} in TSE.}
It is noteworthy that the improvement of TSE dose not just come from the increased parameters by its learners. To see this, we introduce a variant of TSE, \textit{i.e.}, \textbf{TSE-wo-SEO}, 
which adopts a series of ordered learners for consecutive frames without the \textit{saliency erasing operation}. As shown in Tab.~\ref{tab1}, TSE-wo-SEO brings only a small improvement over the baseline, indicating that the visual features captured by different learners are almost the same without the \textit{saliency erasing operation}. While TSE performs significantly better than TSE-wo-SEO, which validates the powerful capability of the \textit{saliency erasing operation} to force different learners to focus on diverse image parts so as to discover integral visual features. Overall, we can see that the improvement of TSE mainly comes from the \textit{saliency erasing operation} rather than the increased parameters.

\begin{table}[t]
\caption{Impact of TSE hyper-parameters on MARS}
\centering
\small
\subfloat[The number of ordered learners]{
\begin{tabular}{l |c  c | c c }
\hline 
$N$  & GFLOPs  & Params & mAP & top-1 \\ 
\hline
1 (base.) &16.246 &23.5M  &79.6 &86.8\\
\hline
2 &16.251 &29.9M & \textbf{82.5} &88.2\\  
3 &16.252  &34.5M &82.4 & \textbf{88.4}\\
4  &16.253  &38.9M &81.0 &87.0\\
\hline
\end{tabular}
\label{tab2b}
}~~~~
\subfloat[Erased block height $h_e$]{
\begin{tabular}{L{0.5cm} |C{1.4cm} C{1.4cm}}
\hline 
$h_e$ & mAP & top-1 \\ 
\hline
2 &81.8 &87.2\\
3 &\textbf{82.5} &\textbf{88.2}\\
4&82.0 &87.4\\
5 &81.7 &87.2 \\
\hline
\end{tabular}
\label{tab2a}
}
\label{tab2}
\end{table}
\textbf{TSE \textit{w.r.t} Number of Ordered Learners.} As shown in Fig.~\ref{TSE} (a), TSE contains $N$ ordered learners that mine complementary parts for $N$ consecutive frames. Tab.~\ref{tab2} (a) studies the impact of the number of learners $N$ on the model base.+TSE. We can observe that the performance increases as more learners are considered to mine complementary parts. However, the performance drops largely when $N$ reaches to $4$. 
In this case, most discriminative parts have been erased in the last frame of the input segment, the fourth learner has to activate non-discriminative regions, \textit{e.g.}, the background, which corrupts the final video representation. Considering the model complexity, we set $N$ to 2 in our work.

\textbf{TSE \textit{w.r.t} Erased Block Size.}
In our network, TSE is applied for the frame-level feature map with size $16\times 8$. Because of the spatial structure of the pedestrian images, we fix the erased width to the width of the feature map  to erase entire rows of the feature map. Tab.~\ref{tab2} (b) studies the impact of the erased height $h_e$ on the performance of TSE. We observe the best performance when $h_e=3$, and it becomes worse when the erased height is larger or smaller. We can conclude: 1) too small erased size cannot effectively encourage the current learner to discover the complementary parts;  2) too large erased size force the current learner to activate non-discriminative image parts, \textit{e.g.}, background.

\textbf{Effectiveness of TSB.} We further assess the effectiveness of TSB module by adding it to the stage2 of baseline (\textbf{base.+TSB}) in Tab.~\ref{tab1}. TSB individually brings $2.7\%$ mAP and $0.8\%$ top-1 accuracy gains with an extremely small increase in computational complexity. The improvements indicate that it is effective to enhance the feature representation power 
by propagating the salient features among the video frames. When we integrate TSE and TSB modules together to TCLNet, the performance can be further improved by about $1\%$ on mAP and top-1 accuracy. 

\textbf{Efficient Positions to Place TSB.} Tab.~\ref{tab1} compares a single TSB module added to different stages of ResNet50 in our TCLNet. The improvements of an TSB module in stage2 and stage3 are similar, but smaller in stage1. One possible explanation is that stage1 has a big spatial size $64\times 32$ that is not very expressive and sufficient to provide precise semantic information. We also present the results of more TSB modules. In particular, we add a TSB module to stage2 and stage3 of the backbone respectively. However, we observe that adding more TSB modules does not bring improvement, indicating that one TSB module is enough for enhancing the salient features.

\textbf{Complexity Comparisons.} As shown in Tab.~\ref{tab1}. We can observe that TSE and TSB introduce negligible computational overhead. In particular, TCLNet requires $16.259$ GFLOPs, corresponding to only $0.08\%$ relative increase over original model (base.). The increased computation cost mainly comes from the correlation map of TSE and probability map of TSB, which can be worked out by matrix multiplications thus occupy little time in GPU libraries. TCLNet introduces $6.4$M parameters to baseline, which mainly come from the learners of TSE. Noting that only extending the baseline with a series of learners (base.+TES-wo-SEO) brings marginal improvement, showing the improvement of TCLNet is not just because of the added parameters.

\textbf{Single shot reID.} TSE is easy to generalize to single shot reID, where different learners discover diverse visual cues for the input image. We compare ResNet50 and ResNet50+TSE on Market1501 dataset~\cite{Market1501}. ResNet50+TSE outperforms ResNet50 by $3.2\%$ mAP ($85.3\%$/$80.3\%$), which indicates the good generality of our method.

\subsection{Comparison with Related Approaches}
\begin{table}[t]
\caption{Compare TSB and TSE modules with related approaches on MARS}
\centering
\small
\subfloat[TSE \textit{vs.} other erasing methods]{
\begin{tabular}{l|c c }
\hline 
Models & mAP & top-1  \\
\hline
base. &79.6 &86.8  \\
\hline
base.+DropBlock~\cite{ghiasi2018dropblock} &79.8 &86.9\\
base.+RE~\cite{zhong2017random} &81.5 &86.6 \\
\hline
base.+TSE &\textbf{82.5} &\textbf{88.2} \\
base.+TSE+RE &\textbf{82.9} &88.1 \\
\hline
\end{tabular}
\label{tab3a}
}~~~~
\subfloat[TSB \textit{vs.} other feature propagation methods]{
\begin{tabular}{l|c c| cc }
\hline 
Models & GFLOPs  & Params & mAP & top-1  \\
\hline
base.+3D~\cite{3D-convolution} &22.756 &33.7M &80.0 &86.1\\
base.+NL~\cite{non-local} &21.615 &25.6M &\textbf{82.4} &\textbf{87.6}\\
base.+TSB &16.254 &23.5M &82.3 &\textbf{87.6}\\
\hline
base.+TSE+NL &21.619 &32.1M &82.6 &87.6\\
TCLNet &16.259 &29.9M &\textbf{83.0} &\textbf{88.8}\\
\hline
\end{tabular}
\label{tab3b}
}
\label{tab3}
\end{table}

\textbf{Comparison TSE with Other Erasing Strategies.} Tab.~\ref{tab3} (a) compares various erasing methods that are applied to baseline. DropBlock~\cite{ghiasi2018dropblock} \textit{randomly} drops a contiguous regions of the convolution features during \textit{training} to \textit{prevent overfitting}. 
As shown in Tab.~\ref{tab3} (a), DropBlock brings marginal improvements while TSE significantly outperforms DropBlock by $2.7\%$ mAP and $1.3\%$ top-1 accuracy. The significant improvements show that it is more efficient to use our erasing strategy to extract complementary features for consecutive frames in video reID tasks. Random Erasing (RE)~\cite{zhong2017random} randomly erases a rectangle region of the \textit{input images} during training, which is a widely used data augmentation technique. As shown in Tab.~\ref{tab3} (a), TSE still outperforms RE. Furthermore, as a data augmentation technique, RE is compatible with our method, in which the performance can be furthered lifted $0.4\%$ mAP when combining with RE.

\textbf{Comparison TSB with Other Feature Propagation Strategies.} Tab.~\ref{tab3} (b) compares TSB with other feature propagation strategies. We can see that our TSB significantly outperforms 3D convolution. We argue that 3D convolution is prone to the corruption of low quality frames and hard to optimize because of the large parameter overhead. Compared with non-local (NL)~\cite{non-local} method, our TSB can achieve comparable performance under less computation budge and model size. More importantly, TSB is more effective to combine with TSE, where TCLNet (TSE+TSB) outperforms base.+TSE+NL by $1.2\%$ top-1 accuracy. We argue that TSB only propagates the \textit{salient information} among video frames which is more complementary to TSE module.

\subsection{Visualization Analysis}
\begin{figure}[t]
\begin{center}
%includegraphics[height=6.5cm]{eijkel2}
\centerline{\includegraphics[width=0.8\textwidth]{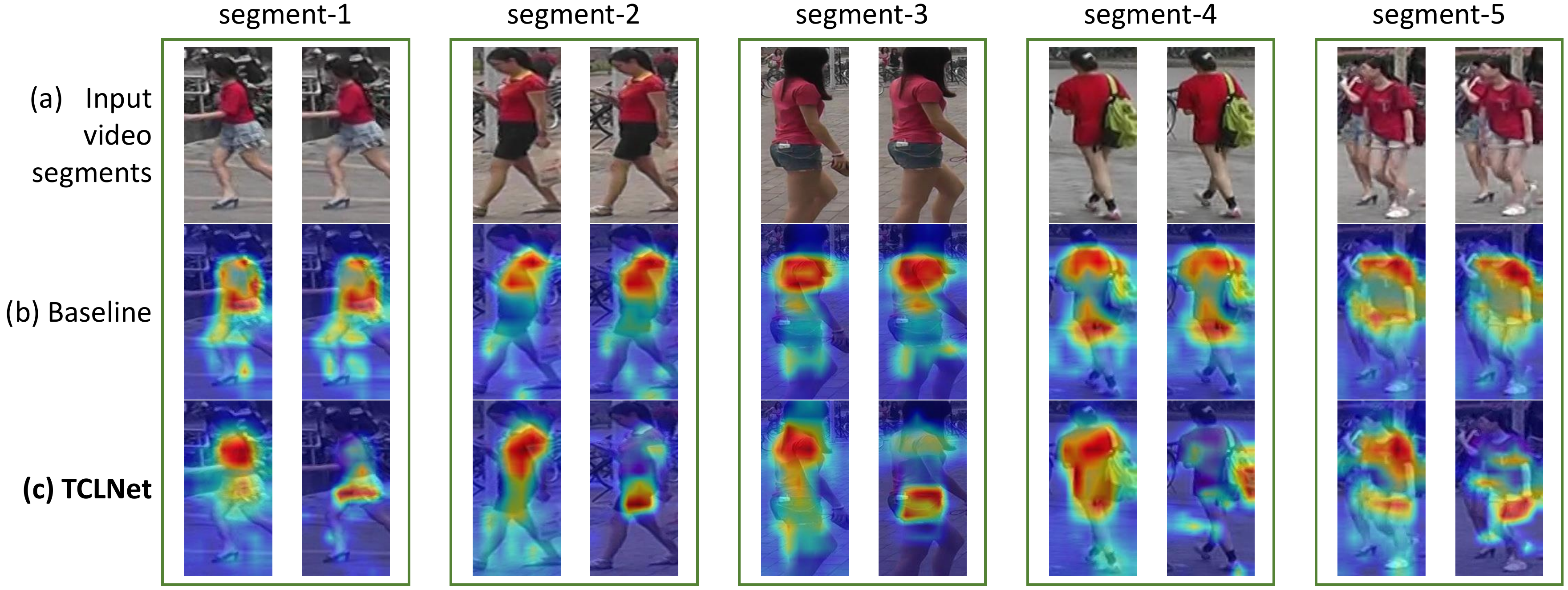}}
\caption{Feature map visualization of baseline and TCLNet}
\label{vis1}
\end{center}
\end{figure}

\textbf{Visualization of Feature Maps.}
For qualitative analysis, we compare the visualization results of feature maps extracted by baseline and TCLNet for some input video segments. As shown in Fig.~\ref{vis1} (b), the features of baseline only pay attention to some local regions, \textit{i.e.}, the red T shirts, which are difficult to distinguish the different pedestrians in Fig.~\ref{vis1} (a). Instead, TCLNet is able to mine complementary parts for consecutive frames. As shown in Fig.~\ref{vis1} (c),  for a video segment consisting of two consecutive frames, the feature of the first frame learned by the learner $L_1$ is most related to the red T shirts, while the learner $L_2$ activates the lower body for the second frame. With the complementary parts mined by learner $L_2$, different persons with similar local appearances become distinguishable. So the final ranking results can be improved significantly.
\begin{figure}[t]
\begin{center}
%includegraphics[height=6.5cm]{eijkel2}
\centerline{\includegraphics[width=0.9\textwidth]{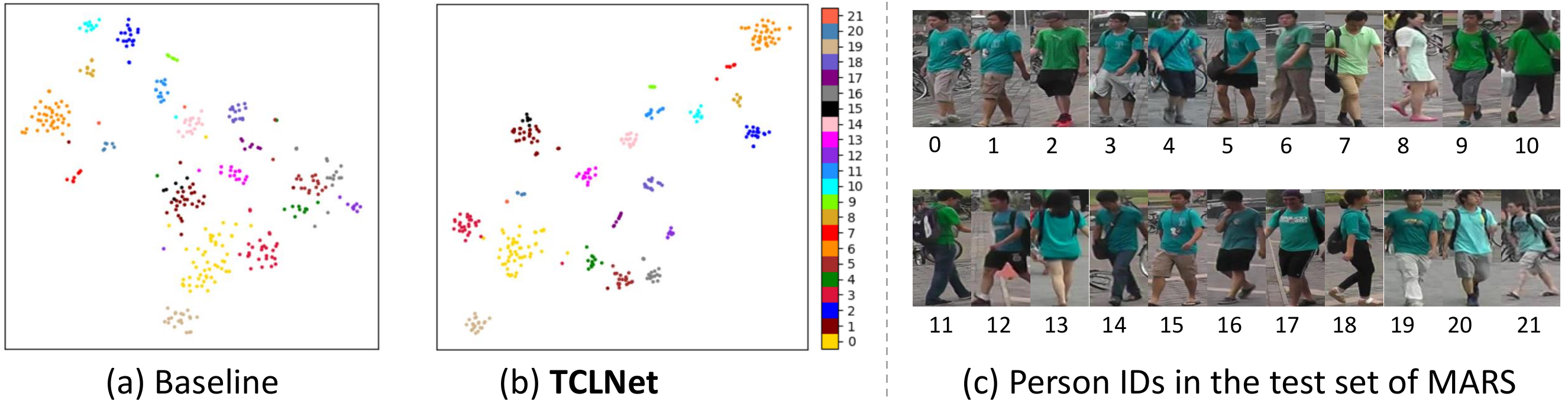}}
\caption{tSNE visualization of feature distribution of baseline and TCLNe on MARS test set. Different colors indicate different identities}
\label{vis2}
\end{center}
\end{figure}

\textbf{Visualization of Feature Distribution.} Furthermore, we choose a number of person IDs with similar appearances from the test set of MARS to visualize the feature distribution by t-SNE~\cite{t-sne}. These pedestrians wear blue shirts with small inter-person variation as shown in Fig.~\ref{vis2} (c). For the baseline that only focuses on local parts, the features belonging to these different identities are staggered. With TSE extracting complementary features and TSB enhancing salient features, the features of different identities extracted by TCLNet become more separable. Specifically, by comparing Fig.~\ref{vis2} (a) and (b), we can observe that for some identities that are hard to be distinguished by baseline, the proposed model can better distinguish them, \textit{e.g.}, the $1^{st}$, $3^{rd}$ and $4^{th}$ identities.

\section{Conclusions}
In this work, we propose a novel Temporal Complementary Learning Network for video person reID. Firstly, we introduce the Temporal Saliency Erasing  module for complementary feature learning of video frames. TSE employs a saliency erasing strategy to progressively discover diverse and complementary visual cues for consecutive frames of a video. 
Furthermore, we propose the Temporal Saliency Boosting module to propagate the salient information among video frames. After propagation, the salient features capture the visual cues of all frames in a video and obtain stronger representational power.  Extensive experiments demonstrate the superiority of our method over current state-of-the-art methods. In the future work, we will improve our method for longer-term temporal modeling. Also, we will combine our method with an efficient strategy to erase the noisy frames for more robust feature representation.

\noindent\textbf{Acknowledgement} 
This work is partially supported by Natural Science Foundation of China (NSFC): 61732004, 61876171 and 61976203.

\bibliographystyle{splncs04}
\bibliography{egbib}
\end{document}